\documentclass[letterpaper, 10 pt, conference]{ieeeconf}  
\usepackage{amsmath,amsfonts}
\include{pythonlisting}
\usepackage[table,xcdraw]{xcolor}
\usepackage{algorithmic}
\usepackage{algorithm}
\usepackage[algo2e,ruled,vlined]{algorithm2e}
\usepackage{array}%\usepackage[caption=false,font=normalsize,labelfont=sf,textfont=sf]{subfig}
\usepackage{textcomp}
\usepackage{stfloats}
\usepackage{url}
\usepackage{verbatim}
\usepackage{graphicx}
\usepackage{cite}
\usepackage{url} 
\usepackage{soul}
\usepackage{multirow}
\usepackage{nicematrix}
\usepackage{subfigure} 
\usepackage{xcolor}
\usepackage{colortbl}
\usepackage{subcaption}
\usepackage{mathrsfs}

\IEEEoverridecommandlockouts                             
\usepackage[utf8]{inputenc}
\usepackage[T1]{fontenc}

\title{\LARGE \bf
Multi-Agent Reinforcement Learning-based Cooperative Autonomous Driving in Smart Intersections
}

\author{Taoyuan Yu, Kui Wang, Zongdian Li, Tao Yu, and Kei Sakaguchi% <-this % stops a space
}

\begin{document}

\maketitle
\thispagestyle{empty}
\pagestyle{empty}

%%%%%%%%%%%%%%%%%%%%%%%%%%%%%%%%%%%%%%%%%%%%%%%%%%%%%%%%%%%%%%%%%%%%%%%%%%%%%%%%
\begin{abstract}

Unsignalized intersections pose significant safety and efficiency challenges due to complex traffic flows. This paper proposes a novel roadside unit (RSU)-centric cooperative driving system leveraging global perception and vehicle-to-infrastructure (V2I) communication. The core of the system is an RSU-based decision-making module using a two-stage hybrid reinforcement learning (RL) framework. At first, policies are pre-trained offline using conservative Q-learning (CQL) combined with behavior cloning (BC) on collected dataset. Subsequently, these policies are fine-tuned in the simulation using multi-agent proximal policy optimization (MAPPO), aligned with a self-attention mechanism to effectively solve inter-agent dependencies. RSUs perform real-time inference based on the trained models to realize vehicle control via V2I communications. Extensive experiments in CARLA environment demonstrate high effectiveness of the proposed system, by: \textit{(i)} achieving failure rates below 0.03\% in coordinating three connected and autonomous vehicles (CAVs) through complex intersection scenarios, significantly outperforming the traditional Autoware control method, and \textit{(ii)} exhibiting strong robustness across varying numbers of controlled agents and shows promising generalization capabilities on other maps. 

\end{abstract}

%%%%%%%%%%%%%%%%%%%%%%%%%%%%%%%%%%%%%%%%%%%%%%%%%%%%%%%%%%%%%%%%%%%%%%%%%%%%%%%%
\section{Introduction}
Intersection management is regarded as a bottleneck in the development of intelligent transportation systems (ITS), considering the complex and uncertain nature of urban intersections \cite{chu10}. According to statistics from the Federal Highway Administration (FHWA) and the National Highway Traffic Safety Administration (NHTSA), intersection-related fatalities account for a substantial proportion of total traffic accident deaths, especially at unsignalized intersections, which reportedly accounted for 68\% of such fatalities in 2024 \cite{mire21, nhtsa2024estimates}. Unsignalized intersections have become accident hotspots due to blind spots and the lack of clear interaction rules between motor vehicles, non-motorized vehicles, and pedestrians. In such environments, connected and autonomous vehicles (CAVs) should possess highly effective perception, prediction, and coordinated decision-making capabilities to minimize conflicts and ensure safe and smooth driving \cite{chen2024review}.

With the popularization of autonomous vehicles (AVs), mixed traffic scenarios involving AVs and human-driven vehicles (HDVs) are becoming common, thereby introducing novel challenges to traffic participants. Vehicle-to-everything (V2X) communication technologies have emerged as a promising solution to improve roadway efficiency and safety \cite{10920203}, typically including vehicle-to-vehicle (V2V), vehicle-to-infrastructure (V2I), vehicle-to-pedestrian (V2P), and vehicle-to-network (V2N) links \cite{li2023het}. Among these, V2I communication facilitates real-time data exchange between CAVs and smart roadside infrastructure (e.g., RSUs), laying a foundation for constructing cooperative driving systems \cite{suo2023proof}. 

Building upon V2I capabilities, the design of RSU-based cooperative systems has appeared as an attractive research topic for both academia and industry in recent years \cite{alemayehu2025testing, sana2023autonomous}. Research on optimizing traffic flow at intersections has explored various methods. Traditional methods often rely on model-based optimization or game-theoretic frameworks to allocate right-of-way and manage traffic, achieving progress in reducing delays under certain conditions \cite{9681232, gallo2024combined, Ghadi_2024}. However, these methods typically lack the adaptability required to effectively handle the high complexity and uncertainty inherent in dynamic real-world traffic scenarios. To address these limitations, multi-agent reinforcement learning (MARL) has emerged as a promising alternative, with research investigating hierarchical structures or integrating perception modules to enhance coordination \cite{shi, tagha, zhang2021trajectory}. Nevertheless, many existing MARL applications in this domain tend to treat all vehicles uniformly, often overlooking the critical need for distinct policies tailored to specific driving roles (e.g., turning left, going straight, turning right) and failing to fully capture the complex interaction dynamics.

Further advancements aim to enhance MARL's capabilities in complex environments. Self-attention mechanisms have been incorporated to dynamically model inter-agent dependencies, potentially improving generalization and decision efficiency \cite{iqbal2019actor, 10900364}. However, the effectiveness validation of adopting self-attention in MARL is still unexplored, particularly its adaptability to dynamically varying numbers of interacting vehicles within realistic transportation contexts like unsignalized intersections. Similarly, hybrid offline-online RL frameworks offer potential benefits by leveraging real-world data for safer and more efficient policy learning \cite{NEURIPS2022_01d78b29, 10611197, wang2019cooperative, wang2021digital}. However, the practical implementation and demonstrated effectiveness of these hybrid approaches in coordinating multiple cooperative vehicles through complex and real-world intersection scenarios still require deeper investigation. Therefore, there exists a research gap in developing and validating an integrated MARL framework capable of robustly and efficiently coordinating vehicles with diverse intentions within the complex dynamics of intersections.

To address such challenges, we propose an innovative RSU-centric intelligent management system for unsignalized intersections. Utilizing bird-eye-view (BEV) perception from RSU-mounted LiDAR, the system employs a centralized MARL decision module featuring role-specific policy networks integrated with a self-attention mechanism, allowing for dynamic modeling of interactions between distinct driving roles and flexible adaptation to varying numbers of vehicle participants. The policy networks are developed using a two-stage hybrid learning approach, involving offline pre-training on the collected dataset, followed by online fine-tuning in the simulation environment. The proposed system demonstrates significant advantages in adaptability, generalization, safety, and efficiency, while also reducing model deployment complexity and computational demands. The key contributions of this research are as follows:

\begin{itemize}
    \item A novel hybrid RL framework combining offline pre-training and online fine-tuning techniques to enable cooperative driving for CAVs at unsignalized intersections.
    \item Development of personalized policy networks tailored to distinct driving roles (e.g., left-turn, straight, right-turn) at intersections.
    \item Integration of a self-attention mechanism into role-based MARL to enhance policy adaptability to varying vehicle numbers and model dynamic interactions.
    \item Demonstration of the model's generalization capability and rapid adaptability across diverse unsignalized intersection scenarios.
\end{itemize}

The remainder of the paper is structured as follows: Section II presents the overall architecture of the RSU-CAVs cooperative system. Section III describes the proposed algorithm in detail. Section IV demonstrates experiment results. Finally, Section V summarizes the paper and outlines future research directions.

\begin{figure*}[t]
     \centerline{\includegraphics[width=0.8 \textwidth]{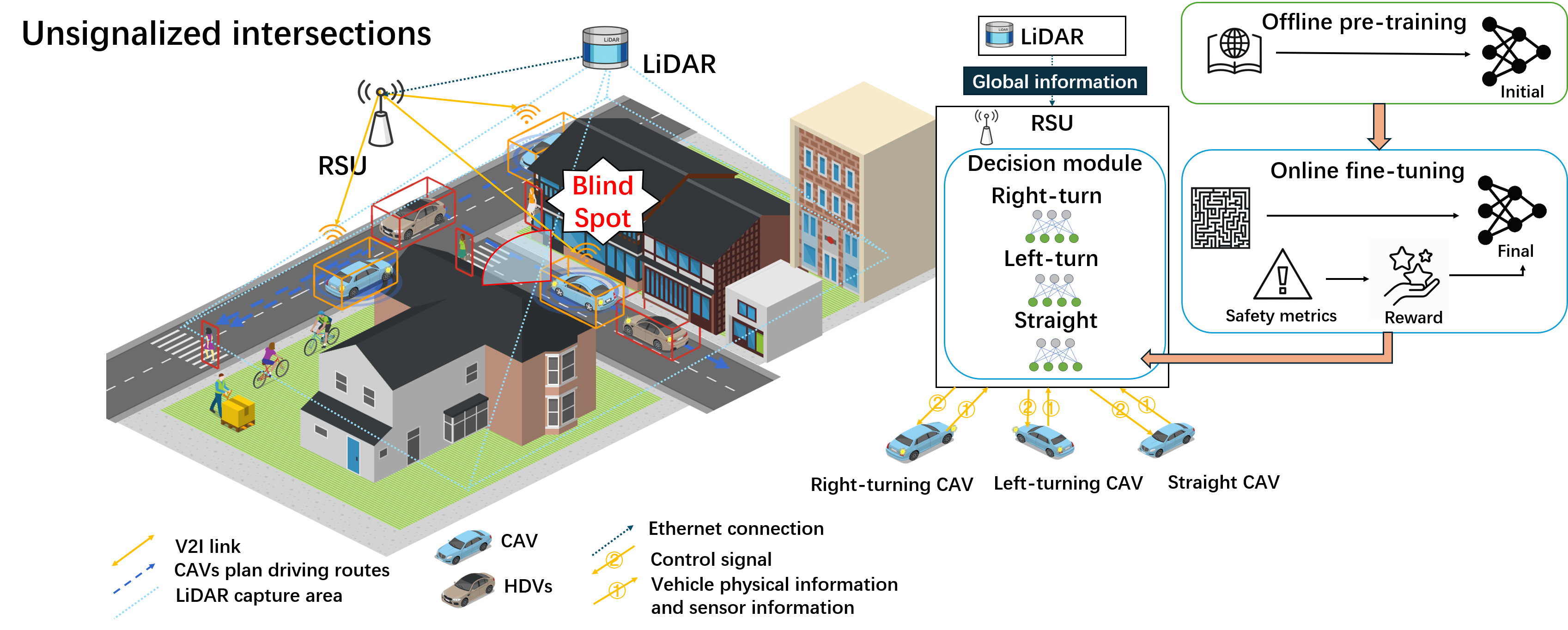}}
     \caption{High-level system design of the RSU-CAVs cooperative system}
     \label{fig: sysd}
\end{figure*}

\section{RSU-CAVs Cooperative System}
The overall system architecture of the proposed RSU-CAV cooperative framework is illustrated in Fig.~\ref{fig: sysd}. This system employs an RSU, equipped with sensors like LiDAR, for comprehensive intersection monitoring \cite{wang2024smdt} and centralized decision-making for multiple CAVs at an unsignalized intersection. Leveraging the RSU's BEV perception, this centralized method overcomes the inherent limitations of individual vehicle perception, providing a global understanding of the traffic situation essential. This contrasts with individualistic methods, where each vehicle optimizes only its own goals, which can lead to competitive standoffs, inefficient gridlocks, or unsafe maneuvers in unsignalized interactions. Instead, our framework enhances collective safety and overall traffic throughput by resolving conflicts harmoniously.

To effectively manage the intersection's complexities and uncertainties, the RSU utilizes adaptive decision-making policies developed through an RL-based method. This method begins by employing offline RL to instill foundational driving knowledge and essential interaction behaviors into the policy from collected datasets, establishing a robust and competent initial strategy. Subsequently, these foundational policies undergo targeted refinement using online RL within a simulation environment. This allows the policies to adapt to the intersection's unique dynamic characteristics and optimize performance for the required multi-vehicle cooperative tasks. Compared to learning entirely from scratch, this hybrid RL approach offers the advantages of accelerating learning convergence during the online refinement phase and ultimately yielding coordination policies that are more robust and effective in handling real-world complexities. Furthermore, deploying the computationally intensive analysis and multi-vehicle coordination logic onto the RSU also reduces the processing burden on individual CAVs. This allocation of tasks improves the system's real-time responsiveness and simplifies requirements for the CAVs \cite{wang2023dtad}. 

As CAVs approach the intersection, they maintain continuous information exchange with the RSU. Based on real-time traffic data and the CAVs' approach trajectories, the RSU determines each vehicle's driving role (e.g., left-turn, straight, right-turn). Subsequently, leveraging its pre-loaded and role-based strategy networks within the centralized decision module, the RSU computes vehicle control signals, including throttle input, braking force, and steering angles. These control commands are transmitted in real-time to the corresponding CAVs through V2I communication, enabling direct command execution. Concurrently, the RSU continuously monitors the comprehensive real-time traffic conditions at the intersection, including the states and predicted movements of CAVs, observed HDVs, and pedestrians, alongside traffic flow smoothness, collision risks, and any abnormal situations. This continuous monitoring provides the necessary real-time inputs for the RSU's decision networks and facilitates ongoing performance evaluation.

\section{Hybrid Reinforcement Learning Framework}
\begin{figure*}[t]
     \centerline{\includegraphics[width=1.0 \textwidth]{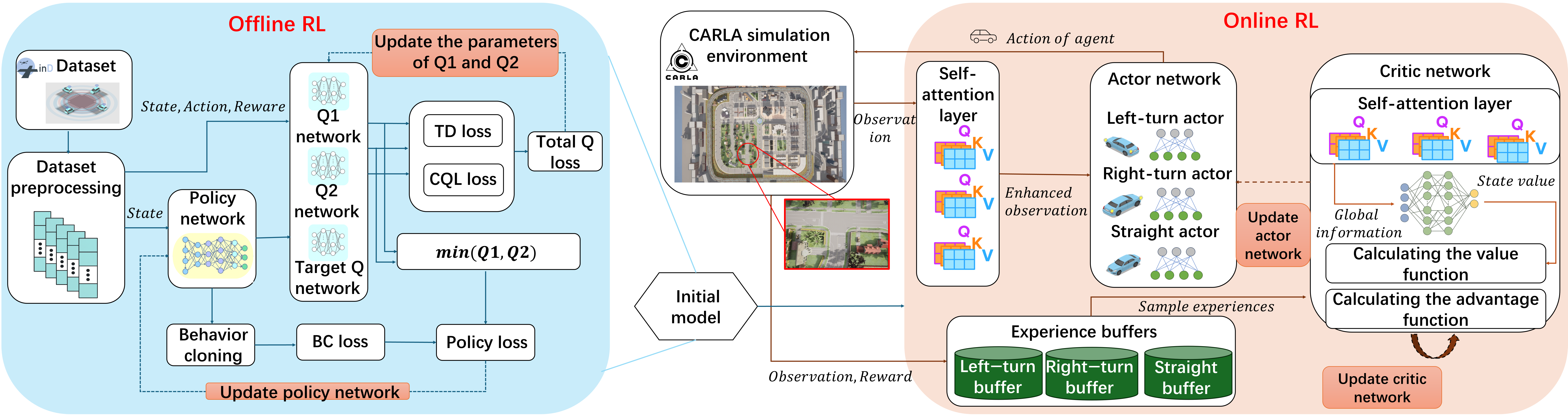}}
     \caption{Offline-online hybrid RL algorithm framework design}
     \label{fig:af}
\end{figure*}

As shown in Fig.~\ref{fig:af}, we propose a two-stage learning framework to develop effective cooperative driving strategies for complex unsignalized intersections. This approach first employs offline pre-training on the collected dataset, using offline RL to safely learn foundational driving skills and traffic priors. Subsequently, online fine-tuning within the CARLA simulator \cite{Dosovitskiy17} allows agents to adapt to specific environment dynamics. This synergistic framework combines the safety and data efficiency of offline learning with the adaptability and performance optimization capabilities of online interaction. The specific methodologies for offline pre-training and online fine-tuning are discussed in this section.

% \subsection{State Space}

% In the simulation environment, we define the state space $\mathcal{S}$, which encompasses all traffic participants monitored and coordinated by RSU within the intersection. At time $t$, a specific state $s(t) \in \mathcal{S}$ can be structured as:
% \begin{equation}
% \begin{aligned}
% s(t) = [
%   s_{\text{cav},1}, \ldots, s_{\text{cav},N_{\text{cav}}},
% & s_{\text{bg},1}, \ldots, s_{\text{bg},M_{\text{bg}}}, \\
% & s_{\text{ped},1}, \ldots, s_{\text{ped},K_{\text{ped}}}
% ]
% \end{aligned}
% \end{equation}

% Here, $N_{\text{cav}}$ denotes the number of agents controlled by the RSU. The $M_{\text{bg}}$ background vehicles and $K_{\text{ped}}$ pedestrians denotes the potential interaction risk. The state of an individual traffic participant $k$ (which can be a CAV, a background vehicle, or a pedestrian) within this overall state vector is represented by $s_k$:

% \begin{equation} \label{eq:individual_state_revised}
% s_k(t) = [\text{type}k, x_k, y_k, \theta_k, v_{x,k}, v_{y,k}, a_{x,k}, a_{y,k}]
% \end{equation}

% where $\text{type}_k$ denotes the participant type, $x_k, y_k$ denotes its global coordinates, $\theta_k$ denotes the heading angle (in radians), $v_{x,k}, v_{y,k}$ denotes its global velocity components, and $ a_{x,k}, a_{y,k}$ denotes its global acceleration components.

\subsection{Observation Space}

At each time step $t$,  the state space can be defined $\mathbf{s}(t)$, which encompasses all traffic participants monitored by RSU within the intersection. the RSU utilizes the global information to generate specific perspective information for each CAV, constructing an individual observation vector $\mathbf{o}(t)$. This construction process, denoted conceptually as $\mathbf{o}(t) = \mathcal{O}(\mathbf{s}(t))$, reflects simulated perception and processing limitations, meaning each $\mathbf{o}(t)$ is a partially observable and potentially noisy representation of the true state $\mathbf{s}(t)$. The observation $\mathbf{o}(t)$ is decomposed as:
\begin{equation}
    \mathbf{o}(t) = [\mathbf{o}_{\text{core}}, \mathbf{o}_{\text{veh}}, \mathbf{o}_{\text{ped}}, \mathbf{o}_{\text{role}}, \mathbf{o}_{\text{ctx}}],
\end{equation}
where $\mathbf{o}_{\text{core}}$ denotes the ego vehicle features including speed magnitude, global position, heading angle, distance to intersection center, and junction occupancy status; $\mathbf{o}_{\text{veh}}$ denotes relative positions and velocities of nearby vehicles within a predefined ranged; $\mathbf{o}_{\text{ped}}$ denotes information on nearby pedestrians, including detection flags, distance, and relative angle; $\mathbf{o}_{\text{role}}$ denotes the one-hot encoding of the agent’s role (e.g., left-turn, straight, or right-turn); and $\mathbf{o}_{\text{ctx}}$ denotes scenario-level identifiers used to curriculum learning.

\subsection{Action Space}

Throughout the learning framework, we define a unified two-dimensional continuous action space $\mathcal{A}$ for the vehicle. It is structured as:
\begin{equation}
\mathbf{a}(t) = [\mathbf{a}_{\text{acc}}, \mathbf{a}_{\text{steer}}] \in \mathbb{R}^2
\end{equation}
where $\mathbf{a}_{\text{acc}}$ denotes the longitudinal acceleration and $\mathbf{a}_{\text{steer}}$ denotes the steering angular velocity. It is important to note that during the offline pre-training phase, as ground-truth control signals are unavailable in the source data, the action $\mathbf{a}(t)$ in the dataset is estimated by analyzing the state transitions between consecutive changes in velocity and heading. In the online fine-tuning phase, the policy network directly outputs two-dimensional action.

\subsection{Reward Function}

To effectively guide the agent in learning desired cooperative driving behaviors during complex online interactions, we design a structured reward function $\mathcal{R}_{\text{online}}(\mathbf{s}(t), \mathbf{a}(t), \mathbf{s}({t+1}))$ to translate high-level objectives into real-time feedback. The overall reward $r(t)$ is structured as:
\begin{equation}
    r(t) = \sum w_k r_k(\mathbf{s}(t), \mathbf{a}(t), \mathbf{s}(t+1)),
\end{equation}
where $r_i$ denotes individual reward components and $w_i$ denotes the corresponding weights. The reward terms include:
\begin{equation}
\begin{aligned}
r_i \in \{ 
    & r_{\text{safety}},\ r_{\text{eff}},\ r_{\text{comfort}}, \\
    & r_{\text{task}},\ r_{\text{yield}},\ r_{\text{coop}},\ r_{\text{penalty}} 
\}
\end{aligned}
\end{equation}
where $r_{\text{safety}}$ denotes the penalty for hazardous behavior based on metrics like minimum time-to-collision (TTC) and distance to nearby vehicles or pedestrians; $r_{\text{eff}}$ denotes the efficiency that encourages maintaining a reasonable speed that is compatible with traffic flow; $r_{\text{comfort}}$ denotes the penalty for large acceleration changes; $r_{\text{task}}$ denotes the reward for all agents to cooperatively reach the navigation target; $r_{\text{yield}}$ and $r_{\text{coop}}$ denotes the rewards for promoting compliance with traffic rules and cooperation; and $r_{\text{penalty}}$ denotes a severe penalty imposed on events like collisions or timeouts. Each term is scaled by its corresponding weight $w_k$, where $w_{\text{safety}}$ and $w_{\text{penalty}}$ are typically assigned larger values due to their critical safety implications.

\subsection{Offline Pre-training: Networks and Algorithm}

The primary goal of the offline pre-training phase is to provide a high-quality initialization for the subsequent online fine-tuning stage. In this stage, we train models independently for each driving role (left-turn, straight, right-turn) to incorporate role-specific prior knowledge. We first partition the InD dataset \cite{inDdataset} based on vehicle intentions to create subsets $\mathcal{D}_{\text{role}}$.

For each subset, we apply an offline reinforcement learning algorithm that combines CQL with BC \cite{NEURIPS2020_0d2b2061}  \cite{NIPS1988_812b4ba2}. The algorithm is implemented using an actor-critic framework to learn effectively from fixed datasets while mitigating distributional shift and imitating expert behavior.

To stabilize learning and reduce Q-value overestimation, the critic uses twin Q-networks $Q_{\theta_{i,1}}$, $Q_{\theta_{i,2}}$ with target networks. Each Q-network is trained with the following objective:
\begin{equation}
\begin{aligned}
    L_Q(\theta_{i,j}) = & \mathbb{E}_{(\mathbf{o},\mathbf{a},r,\mathbf{o'}) \sim \mathcal{D}_{\text{role}=i}} \left[ \frac{1}{2}(Q_{\theta_{i,j}}(\mathbf{o},\mathbf{a}) - y)^2 \right] \\
    & + \alpha_{\text{CQL}} L_{\text{CQL\_reg}}(\theta_{i,j})
\end{aligned}
\end{equation}
Here, $y = r + \gamma (1 - d) \min_j Q_{\theta_{i,j}'}(\mathbf{o}', \pi_{\phi_i}(\mathbf{o}'))$ denotes the TD target computed using target Q networks and the current policy.

The policy network $\pi_{\phi_i}$ is trained by minimizing the BC loss along with maximizing the expected conservative Q-value:
\begin{equation}
\begin{aligned}
    L_{\pi}(\phi_i) = & \mathbb{E}_{\mathbf{o} \sim \mathcal{D}_{\text{role}=i}} \left[ - \min_{j=1,2} Q_{\theta_{i,j}}(\mathbf{o}, \pi_{\phi_i}(\mathbf{o})) \right] \\
    & + \lambda_{\text{BC}} \mathbb{E}_{(\mathbf{o},\mathbf{a}) \sim \mathcal{D}_{\text{role}=i}} \left[ \lVert \pi_{\phi_i}(\mathbf{o}) - \mathbf{a} \rVert^2 \right]
\end{aligned}
\end{equation}
where $\alpha_{\text{CQL}}$ and $\lambda_{\text{BC}}$ denotes hyperparameters controlling the strength of CQL regularization and BC imitation, respectively.

Each role-specific actor $\pi_{\phi_{\text{role}}}$ and critic $Q_{\theta_{\text{role}}}$ are parameterized by multi-layer perceptrons (MLPs), which take normalized state inputs $s_t$ sampled from $\mathcal{D}_{\text{role}}$. The output corresponds to action distribution parameters or state-action values.

Self-attention are not introduced at this stage to focus training on extracting robust role-based patterns using standard MLPs. Training stability is further improved by soft target updates and the adam optimizer. Successful offline training yields a set of pre-trained weights for the role-conditioned actor and critic networks, which are reused during the online fine-tuning phase to accelerate learning and enhance performance.

\subsection{Online Fine-tuning: Networks and Algorithm}

The online fine-tuning phase employs the MAPPO algorithm \cite{yu2021ppo}, selected for its effectiveness in multi-agent coordination. This stage builds upon the offline models by integrating role-specific actor networks ($\pi_{\phi_{\text{left}}}, \pi_{\phi_{\text{straight}}}, \pi_{\phi_{\text{right}}}$) with a shared critic network $V_\psi$.

To improve reasoning over dynamic environments, we augment both actor and critic networks with multi-head self-attention (MHSA). MHSA allows the model to jointly attend to information from different representation subspaces at different positions. The core component is the scaled dot-product attention:
\begin{equation}
\label{eq:attention}
\text{A}(Q, K, V) = \text{softmax}\left(\frac{QK^\top}{\sqrt{d_k}}\right) V
\end{equation}
where $Q$, $K$, and $V$ denote the query, key, and value matrices, and $d_k$ denotes the dimension of the keys. MHSA computes $h$ attention 'heads' in parallel. For each head $i$, the input embedding $E$ is linearly projected using learned weights $W_i^Q$, $W_i^K$, $W_i^V$ to obtain the head's specific query, key, and value:
\begin{equation}
\label{eq:head_i}
\text{h}_i = \text{Attention}(E W_i^Q, E W_i^K, E W_i^V)
\end{equation}
The outputs of the parallel heads are then concatenated and linearly projected using weights $W^O$ to produce the final MHSA output:
\begin{equation}
\label{eq:mhsa}
\text{M}(E) = \text{Concat}(\text{head}_1, \ldots, \text{head}_h) W^O
\end{equation}
Here, $E$ denotes the initial embedded observation features derived from the online observation $\mathbf{o}_t$ via a learnable linear projection layer, and $W_i^Q$, $W_i^K$, $W_i^V$, $W^O$ denote trainable weight matrices.

Online learning proceeds via an interact-learn loop. Agents generate trajectories:
\begin{equation}
\tau = \left\{(\mathbf{o}_t, \mathbf{a}_t, r_{t+1}, V_\psi(\mathbf{o}_t), \log \pi_{\phi_{\text{role}}}(\mathbf{a}_t \mid \mathbf{o}_t))\right\}_{t=0}^{T}
\end{equation}

Advantage estimates and returns are computed using generalized advantage estimation (GAE):
\begin{equation}
\hat{A}_t^{\text{GAE}} = \sum_{l=0}^{T-t-1} (\gamma \lambda)^l \delta_{t+l}, \quad \delta_t = r_{t+1} + \gamma V_\psi(\mathbf{o}_{t+1}) - V_\psi(\mathbf{o}_t)
\end{equation}
\begin{equation}
\hat{R}_t = \hat{A}_t^{\text{GAE}} + V_\psi(\mathbf{o}_t)
\end{equation}

To enhance data efficiency, we adopt prioritized experience replay (PER). Each transition $t$ is assigned a priority $p_t$ proportional to its absolute TD error $|\delta_t|$, and sampled with probability $P(t) \propto p_t$. To correct the bias introduced by this non-uniform sampling, importance sampling (IS) weights are applied:
\begin{equation}
w_t = \left( \frac{1}{B \cdot P(t)} \right)^\beta
\end{equation}
Here, $B$ denotes the size of the replay buffer, and $\beta$ denotes an exponent that controls the amount of importance sampling correction.

The shared critic is updated to minimize the weighted value loss:
\begin{equation}
L^{VF}(\psi) = \mathbb{E}_{t \sim \text{PER}} \left[ w_t (V_\psi(\mathbf{o}_t) - \hat{R}_t)^2 \right]
\end{equation}
where $\hat{R}_t$ denotes the target return calculated in Eq.(12).

Each role-specific actor $\pi_{\phi_{\text{role}}}$ is trained using the following weighted objective, which includes the PPO clipped surrogate loss and an entropy bonus $S[\cdot]$:
\begin{equation}
\begin{aligned}
L^{\text{CLIP+S}}(\phi_{\text{role}}) = \mathbb{E}_{t \sim \text{PER}} \Big[ w_t \big(
& - L_t^{\text{CLIP}}(\phi_{\text{role}}) \\
& - c_2 \cdot S[\pi_{\phi_{\text{role}}}](\mathbf{o}_t) \big) \Big]
\end{aligned}
\end{equation}

The PPO surrogate loss $L_t^{\text{CLIP}}$ is defined as:
\begin{equation}
L_t^{\text{CLIP}} = \min \left( r_t \hat{A}_t, \text{clip}(r_t, 1 - \epsilon, 1 + \epsilon) \hat{A}_t \right)
\end{equation}
where $\epsilon$ denotes the PPO clipping hyperparameter, and $r_t$ denotes the probability ratio between the current policy and the old policy:

\begin{equation}
r_t = \frac{\pi_{\phi_{\text{role}}}(\mathbf{a}_t \mid \mathbf{o}_t)}{\pi_{\phi_{\text{old}}}(\mathbf{a}_t \mid \mathbf{o}_t)}
\end{equation}

We further enhance training with adam optimizer and gradient clipping. These stabilizing techniques enable robust fine-tuning and effective adaptation to dynamic, multi-agent environments.

\section{Experiments and Analysis}
\begin{figure}[t]
\centering
\subfigure[]{
\begin{minipage}[b]{0.4\textwidth}
\includegraphics[width=1\textwidth]{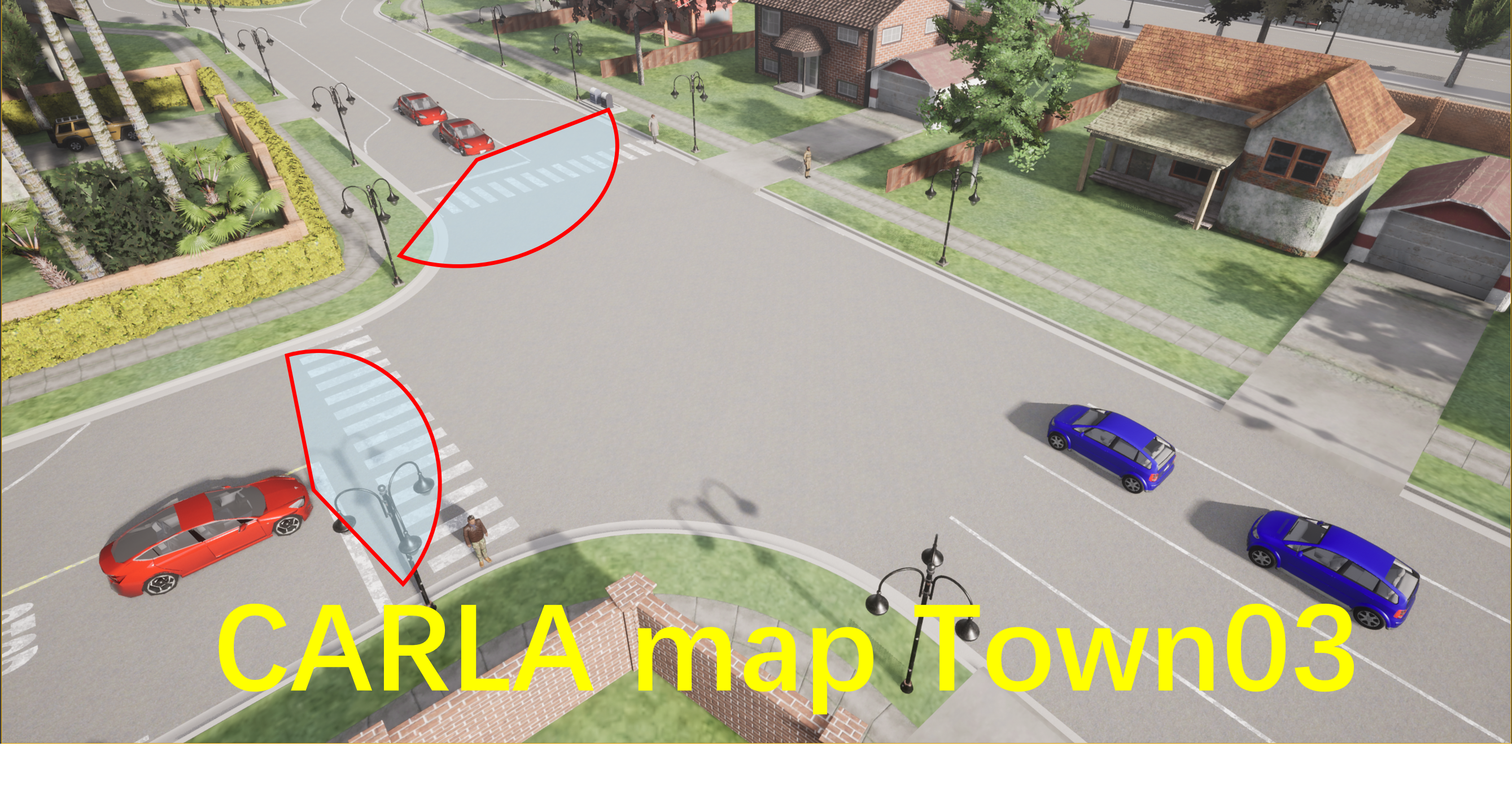} 
\end{minipage}
}
\subfigure[]{
\begin{minipage}[b]{0.4\textwidth}
\includegraphics[width=1\textwidth]{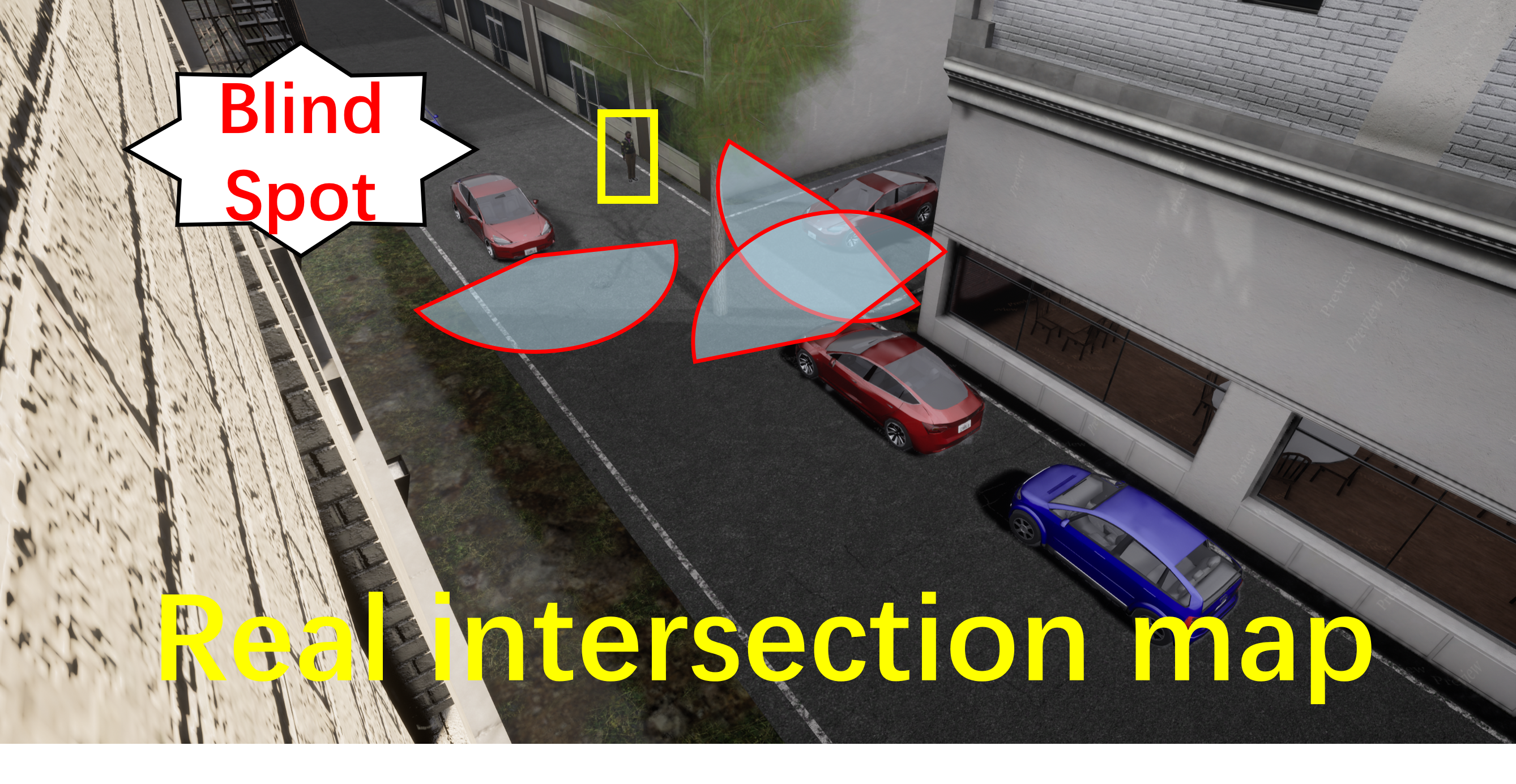} 
\end{minipage}
}
\caption{Experimental scenario and generalization scenario settings (a) CARLA example map, (b) Real intersection map}
\label{fig:tu}
\end{figure}

Experiments were conducted using the CARLA simulator paired with Unreal Engine in synchronous mode. The primary scenario involves one intersection within CARLA's Town03 map (shown in Fig.~\ref{fig:tu}). In this simulation scenario, a total of 5 vehicles are present: a variable number (1 to 3) CAVs, designated "red", are controlled by the proposed system, while the remaining background vehicles, designated "blue", are managed by CARLA's Traffic Manager. Additionally, up to 3 pedestrians are randomly spawned on sidewalks and programmed to cross the road. A real intersection map based on the Institute of Science Tokyo campus was utilized for generalization testing. We assume the RSU possesses BEV perception and performs inference using the decision model obtained after online fine-tuning to determine driving strategies, subsequently sending control signals to the CAVs via simulated V2I communication.

%\begin{figure}[t]
%     \centerline{\includegraphics[width=0.5 \textwidth]{fig/3.png}}
 %    \caption{Experimental scenario and generalization scenario settings}
  %   \label{fig:tu}
%\end{figure}

\subsection{Baselines and Evaluation Metrics}

We used online-only MAPPO as part of an ablation study; we trained a MAPPO agent that undergoes no offline pre-training and starts training directly in CARLA from scratch (with the network structure and algorithm parameters being the same as the online phase of our proposed method), used to compare and evaluate the performance of our proposed algorithm. We will primarily compare them in terms of convergence speed. Additionally, we employ the open-source autonomous driving software stack, Autoware Universe \cite{Autoware} as a representative of traditional autonomous driving systems. In our experiments, Autoware is configured to control a single vehicle navigating the intersection within the identical scenario used for our single-agent RL tests (i.e., 1 Autoware-controlled vehicle, 4 background Traffic Manager vehicles, and 3 pedestrians). Its performance provides a benchmark for comparison against established methods.

\subsection{Offline Pre-training Results}

% \begin{figure*}[t]
% \centering
% \subfigure[]{
% \begin{minipage}[b]{0.3\textwidth}
% \includegraphics[width=1\textwidth]{fig/4.1.png} 
% \end{minipage}
% }
% \subfigure[]{
% \begin{minipage}[b]{0.3\textwidth}
% \includegraphics[width=1\textwidth]{fig/4.2.png} 
% \end{minipage}
% }
% \subfigure[]{
% \begin{minipage}[b]{0.3\textwidth}
% \includegraphics[width=1\textwidth]{fig/4.3.png} 
% \end{minipage}
% }
% \caption{Offline pre-training results (a) Q1-funtion loss, (b) Q2-funtion loss, (c) Reward improvement during training}
% \label{fig:4}
% \end{figure*}

\begin{figure}[t]
     \centerline{\includegraphics[width=0.45 \textwidth]{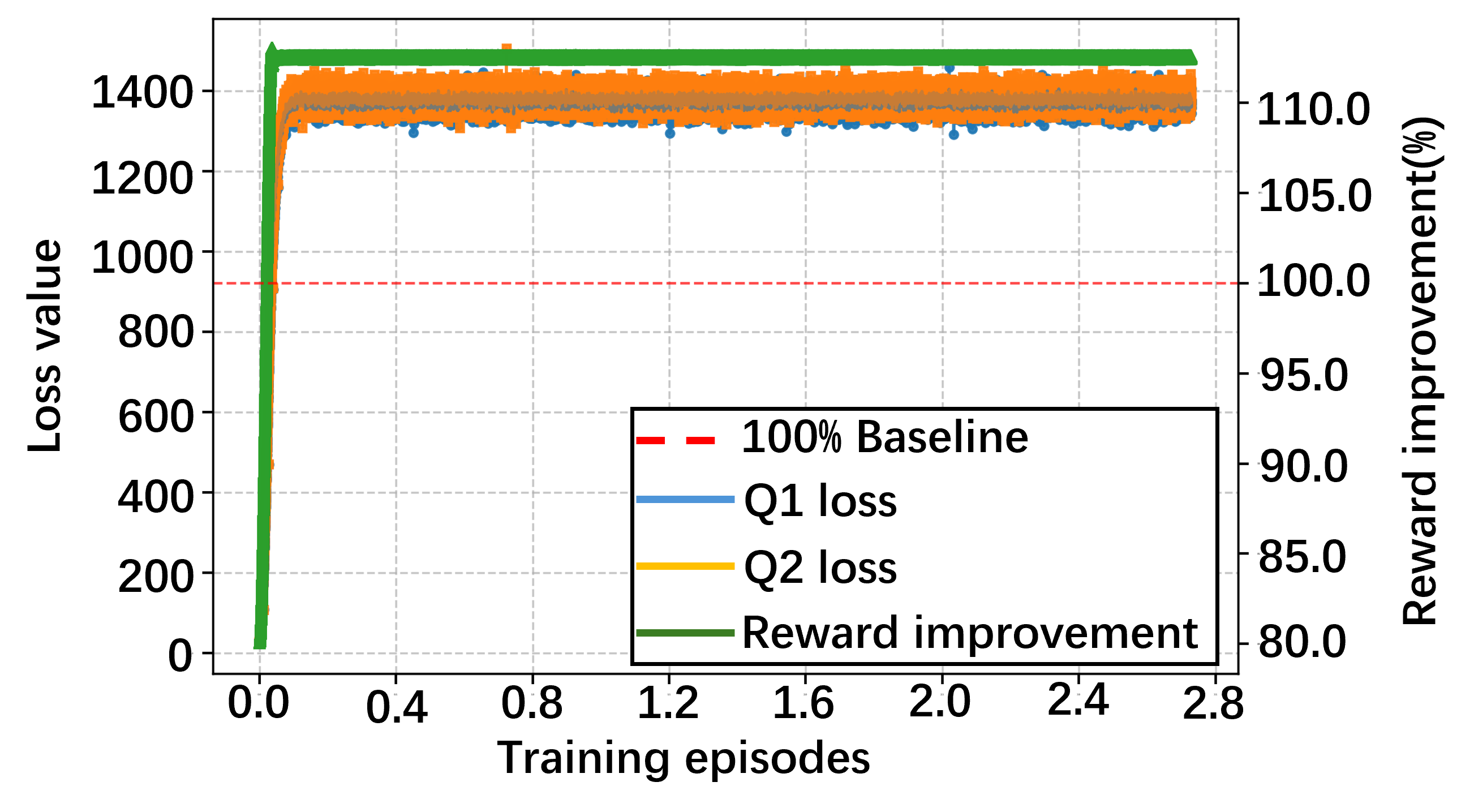}}
     \caption{Offline pre-training results}
     \label{fig:4}
\end{figure}

The purpose of the offline pre-training phase is to extract effective driving priors from the InD dataset, providing a high-quality model initialization for the online phase. As shown in Fig.~\ref{fig:4}, which displays the changes in the critic networks' Q1 and Q2 losses and the reward improvement metric during offline training, we can observe the learning progress. The loss values steadily converge as training progresses, indicating that the critic network effectively learned state-action value relationships from the offline data and that the training process possessed good stability. Furthermore, the reward improvement metric eventually stabilizes around 112\%, exceeding the 100\% baseline. This demonstrates that the policy learned via CQL combined with BC outperforms the average behavior present in the dataset in terms of optimizing the offline reward objective, successfully learning strategies beyond mere imitation, and providing a quality initialization basis for online fine-tuning.

\subsection{Online Training Results}

\begin{figure}[t]
\centering
\subfigure[]{
\begin{minipage}[b]{0.4\textwidth}
\includegraphics[width=1.\textwidth]{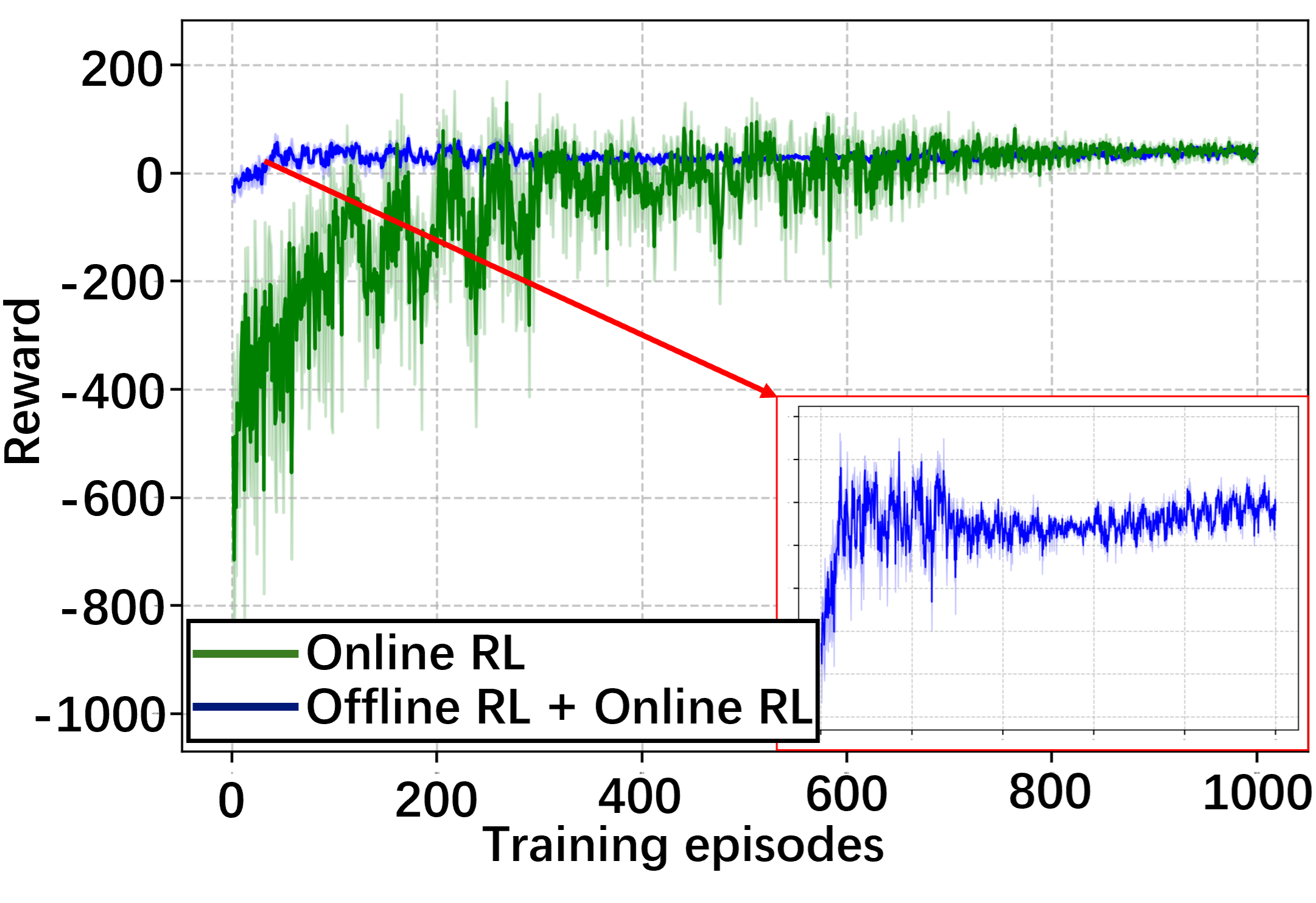} 
\end{minipage}
}
\subfigure[]{
\begin{minipage}[b]{0.4\textwidth}
\includegraphics[width=1\textwidth]{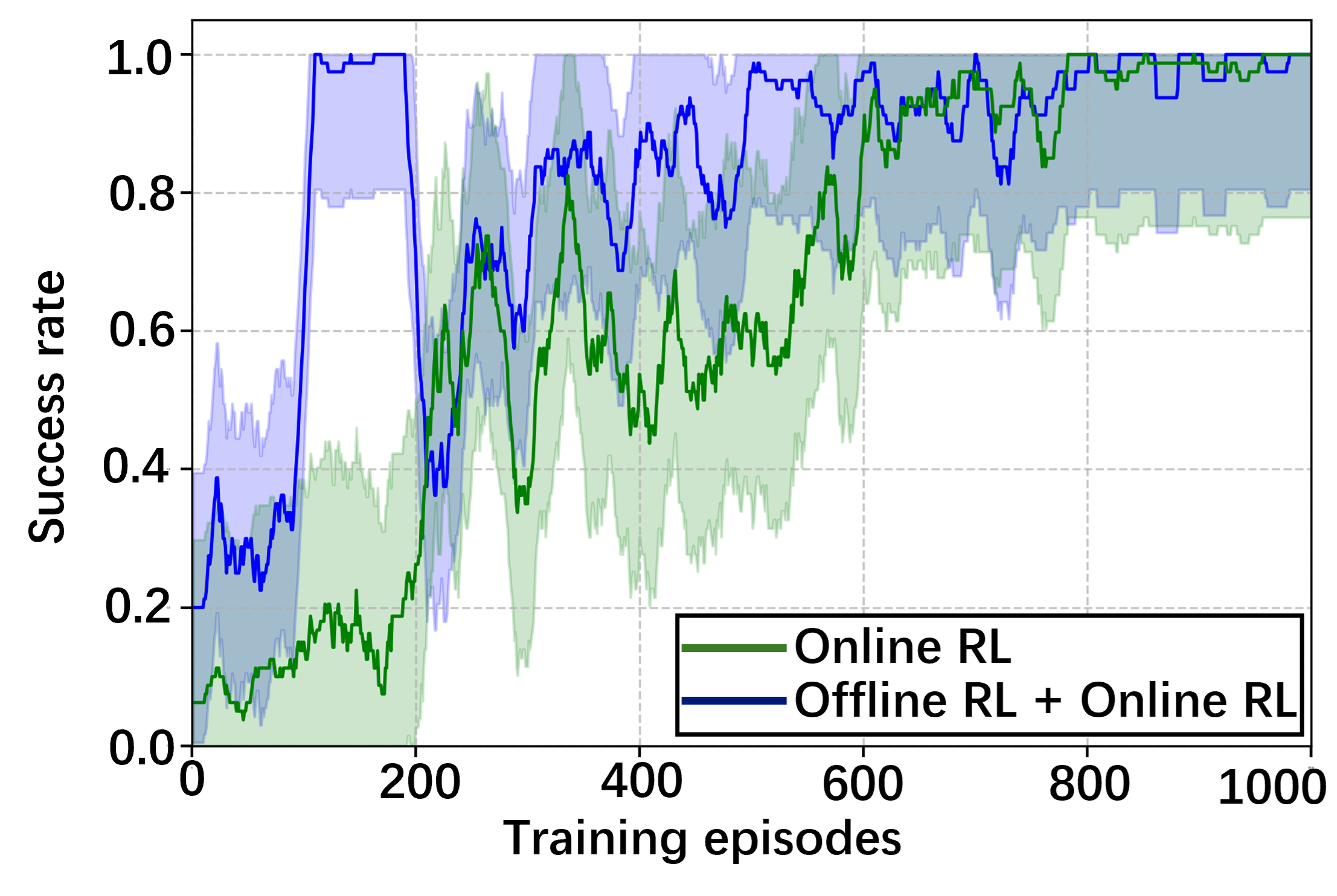} 
\end{minipage}
}
\caption{Comparison of online training w/ and w/o offline RL, (a) reward, (b) success rate.}
\label{fig:5}
\end{figure}

To validate the effectiveness of online fine-tuning and the value of offline pre-training, we compare the training progress of our proposed hybrid method against the online-only baseline. Fig.~\ref{fig:5} presents the convergence curves for both average episode reward and success rate during the external online training phase for both methods.

As shown in Fig.~\ref{fig:5}, our proposed hybrid method, leveraging the pre-trained model, exhibits a much higher initial average episode reward compared to the online-only method starting from scratch. Furthermore, the hybrid approach converges more rapidly to its final reward level, indicating that offline pre-training significantly accelerates the online learning process and contributes to achieving strong final performance. Additionally, the convergence trend for success rate mirrors that of the reward. Our proposed method reaches high success rates considerably faster, whereas the online-only method requires significantly more training episodes to approach similar levels of reliability. This comparison further demonstrates that offline pre-training markedly enhances both the efficiency of the online learning phase and the robustness of the ultimately learned policy.

\subsection{Final Model Performance Evaluation and Generalization Analysis}

\begin{table}[t]
\centering
\caption{Performance comparison summary}
\label{tab:performance_comparison_simple}
\scriptsize % 把字体再缩小一点
\begin{tabular}{|c|c|c|}
\hline
\textbf{Method / Scenario} & \textbf{Failure rate (\%)} & \textbf{Avg. time (s)} \\
\hline
Ours (1 Agent, Town03) & 0.01 & 5.52 \\
Ours (2 Agent, Town03) & 0.03 & 5.49 \\
Ours (3 Agent, Town03) & 0.02 & 5.25 \\
\hline
Autoware (1 Agent, Town03) & 5.31 & 5.77 \\
\hline
Ours (3 Agent, Real Map) & 0.02 & 5.15 \\
\hline
\end{tabular}
\end{table}

\begin{figure}[t]
\centering
\subfigure[]{
\begin{minipage}[b]{0.4\textwidth}
\includegraphics[width=1\textwidth]{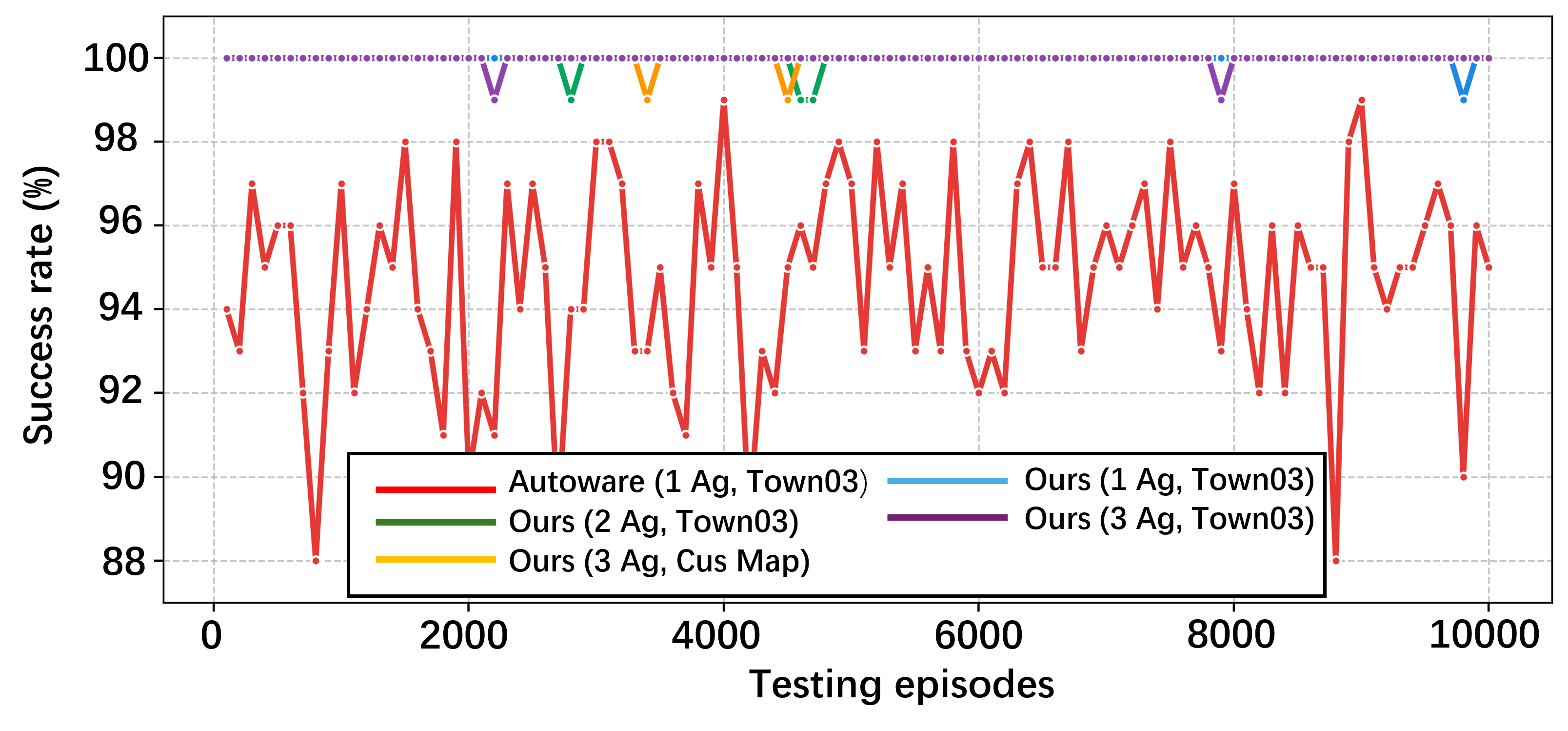} 
\end{minipage}
}
\subfigure[]{
\begin{minipage}[b]{0.4\textwidth}
\includegraphics[width=1\textwidth]{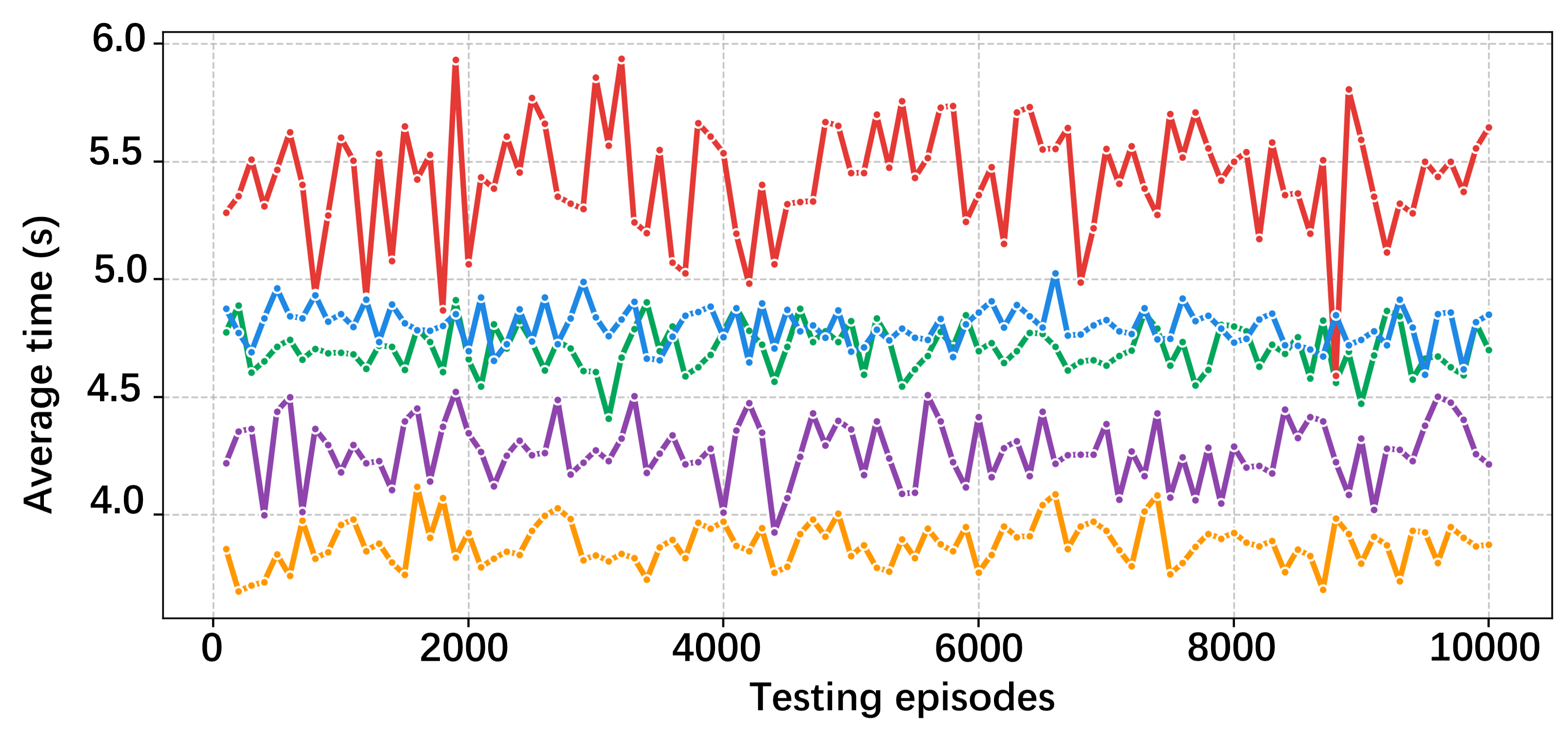} 
\end{minipage}
}
\caption{Final model performance evaluation results (a) success rate by testing episodes, (b) average travel time by testing episodes}
\label{fig:6}
\end{figure}

We evaluated the model through 10,000 performance test episodes on both the Town03 intersection and the custom map, comparing it against baselines. A summary of key performance indicator comparisons is shown in Fig.~\ref{fig:6} and Tab.~\ref{tab:performance_comparison_simple}. The proposed algorithmic model demonstrates high safety and reliability across all test scenarios on the Town03 map. When controlling a single vehicle, the failure rate was 0.01\%. Compared to the 5.31\% failure rate exhibited by the Autoware baseline in the identical single-vehicle scenario, our single-agent controller shows higher performance. %highlighting the superior adaptability and decision-making capabilities of the RL approach in handling complex dynamic interactions.

Notably, despite the significant increase in coordination complexity when scaling from single- to multi-vehicle scenarios, our system did not exhibit a marked decline in success rate. Specifically,  failure rate was 0.03\% in the two-vehicle coordination scenario and decreased to 0.02\% in the three-vehicle coordination scenario. The combination of the RSU's BEV perspective and the self-attention mechanism contributes to this robustness, demonstrating our method's effectiveness in handling complex multi-agent cooperative tasks.

As shown in Fig.~\ref{fig:6} and Tab.~\ref{tab:performance_comparison_simple}, the traffic efficiency results indicate that our method also demonstrated strong performance. The average travel time in the single-vehicle scenario was 5.52 seconds, outperforming the 5.77 seconds recorded by the Autoware baseline. As the number of controlled vehicles increased, the average travel time showed a slight downward trend: 5.49 seconds for the two-vehicle scenario and 5.25 seconds for the three-vehicle scenario. This indicates that the multiple RL agents coordinated by our system formed a highly effective collaborative passage pattern. Their interactions proved more efficient than those with a larger number of background vehicles controlled by the Traffic Manager.

Finally, in the generalization test, the three-vehicle model trained in Town03 was deployed on the real intersection map. The model achieved an extremely low failure rate of 0.02\% and an average travel time of 5.15 seconds in this novel environment. This suggests that the expanded collective field of view inherent in the three-vehicle setup mitigated the impact of individual visual blind spots. Furthermore, as this map featured shorter traversal distance, it enabled our system to operate more smoothly and efficiently. This result strongly validates the excellent generalization capability of the learned policy, showcasing its adaptability to different environmental characteristics and providing a solid foundation for the practical application of the method.

\section{Conclusion and Future Works}
%This research addresses the complex coordination challenges at unsignalized intersections by proposing and validating an RSU-based centralized cooperative driving framework. The framework employs a two-stage method to train the decision model: offline pre-training initializes policies, followed by online fine-tuning in the simulation environment. This model leverages role-specific policies and self-attention mechanisms to handle multi-agent interactions and ultimately deployed to the RSU for real-time inference and control. Extensive experiments demonstrate the method's effectiveness, achieving high success rate and strong coordination robustness in scenarios with up to three controlled vehicles, along with good generalization demonstrated on other map. Future primary work involves the Proof of Concept (PoC) on the Institute of Science Tokyo campus, deploying the RSU system to coordinate two physical CAVs interacting with pedestrians to evaluate performance. Furthermore, exploring the use of neural networks for potentially improved interaction modeling and testing the system's scalability in more complex scenarios are important avenues for future research.

This research addresses the complex coordination challenges at unsignalized intersections by proposing an RSU-based centralized cooperative driving framework. The framework employs a two-stage method to train the decision model: offline pre-training initializes policies, followed by online fine-tuning in the simulation environment. Extensive experiments demonstrate the method's effectiveness, achieving high success rate and strong coordination robustness in scenarios with up to three controlled vehicles. Future primary work involves the proof-of-concept (PoC) experiments to fully validate the system effectiveness in the real world.

\bibliographystyle{IEEEtran.bst}
\bibliography{bibliography}

\end{document}